\documentclass{article} 
\usepackage{iclr2015,times}
\usepackage{hyperref}
\usepackage{url}
\usepackage{graphicx}
\usepackage{bm}
\usepackage{amsmath}
\usepackage{amsfonts}
\usepackage{array}
\usepackage{titlesec}

\titlespacing*{\section}{0pt}{1pt}{1pt}

\def\vtheta{{\bm{\theta}}}

\def\vx{{\bm{x}}}

\title{On Distinguishability Criteria for\\Estimating Generative Models}

\author{
Ian J. Goodfellow \\
Google Inc., Mountain View, CA\\
\texttt{goodfellow@google.com} \\
}

\iclrfinalcopy 
\begin{document}
\maketitle
\begin{abstract}
    \vspace{-.1in}
    Two recently introduced criteria for estimation of generative models
    are both based on a reduction to binary classification.
    Noise-contrastive estimation (NCE) is an estimation procedure in which
    a generative model is trained to be able to distinguish data samples from
    noise samples. Generative adversarial networks (GANs) are pairs of generator and
    discriminator networks, with the generator network learning to generate samples by
    attempting to fool the discriminator network into believing its samples are
    real data.
    Both estimation procedures use the same function to drive learning,
    which naturally raises questions about how they are related
    to each other, as well as whether this function is related to
    maximum likelihood estimation (MLE).
   NCE corresponds to training an internal data model
    belonging to the {\em discriminator} network but using a fixed generator network.
    We show that a variant of NCE, with a dynamic generator network,
    is equivalent to maximum likelihood estimation.
    Since pairing a learned discriminator with an appropriate dynamically selected
    generator recovers MLE, one might expect the reverse to hold for pairing a learned
    generator with a certain discriminator. However, we show that recovering MLE
    for a learned generator
    requires departing from the distinguishability game.
    Specifically:
    (i) The expected gradient of the NCE discriminator can be made to match the expected gradient of
    MLE, if one is allowed to use a non-stationary noise distribution for NCE,
    (ii) No choice of discriminator network can make the expected gradient for the GAN
    generator match that of MLE, and
    (iii) The existing theory does not guarantee that GANs will converge in the non-convex case.
    This suggests that the key next step in GAN research is to determine whether GANs converge,
    and if not, to modify their training algorithm to force convergence.
\vspace{-.05in}
\end{abstract}
\vspace{-.1in}
\section{Introduction}
    Many machine learning applications involve fitting a probability distribution over a vector of observations
    $\vx$. This is often accomplished by specifying a parametric family of probability distributions indexed
    by a parameter vector $\vtheta$.
    The resulting $p_\text{model}(\vx; \vtheta)$ can be approximately
    matched to the reference distribution $p_\text{data}(\vx)$ by using a statistical estimator to find
    a good value of $\vtheta$.

    Maximum likelihood estimation (MLE) is the most popular statistical estimator used to accomplish this task.
    Maximum likelihood estimation works by maximizing the probability of the observed data according to the model.
    For several models of interest, exact MLE is intractable, and must be approximated using techniques such
    as Markov chain Monte Carlo methods. As a result, one popular avenue of research is the design of alternative
    statistical estimators that have lower computation cost.
    
    Two recently proposed methods both employ a binary classifier that attempts to discern whether a given $\vx$
    was drawn from the training data or sampled from a ``generator'' distribution. Both methods are primarily
    driven by a function we call the {\em distinguishability game value function}:

    \[ V(p_c, p_g) = %
        \mathbb{E}_{\vx \sim p_d} \log p_c(y=1 \mid \vx) +
\mathbb{E}_{\vx \sim p_g} \log p_c(y=0 \mid \vx)  \]
    where $p_g(\vx)$ is a distribution representing a source of ``fake'' samples, $p_d(\vx)$ is the distribution
    over the training data (which we can usually access only approximately via an emprical distribution),
    and $p_c(y = 1 \mid \vx)$ is the classifier's probability that $\vx$ comes from $p_d$ rather than $p_g$.

    The distinguishability game value function is (proportional to) the log likelihood of the classifier
    on the binary class labeling problem, where the training examples for the classifier are drawn from a uniform mixture of
    two components: real data labeled with $y=1$ and ``fake'' samples labeled with $y=0$.
    
    Of the two recently proposed methods, the first is {\em noise-contrastive estimation}~\citep{nce}.
    Noise-contrastive estimation (NCE) uses an arbitrary, fixed ``noise'' distribution for $p_g$.
    The goal of NCE is to learn a model $p_m(\vx)$ that is used to define the classifier:
    \begin{equation}
        p_c(y=1\mid \vx) = \frac{ p_m(\vx) } { p_m(\vx) + p_g(\vx) }.
            \label{eq:nce_c}
        \end{equation}
    Learning then proceeds by using a standard optimization algorithm to maximize $V$.

    More recently, the generative adversarial network (GAN) framework introduced a different approach
    using the same value function~\citep{gan}. Here, the goal is to learn $p_g$, and there is no
    explicit $p_m$. Instead, $p_c$ is parameterized directly. Rather than treating $V$ as an objective
    function to maximize, it is used to define a minimax game, with $p_g$ trained to minimize the
    objective and $p_c$ trained to maximize it.

    MLE, NCE, and GANs are all asymptotically consistent, which means that in the limit of 
    infinitely many samples from $p_d$ being available, their criteria each have a unique stationary
    point that corresponds to the learned distribution matching the data distribution\footnote{
    By ``match'' we mean more formally that $p_d(\vx) = p_m(\vx)$ for all $\vx$ except a set of measure zero.
    }.
    For MLE and NCE, this stationary point is a global maximum of their  objective function, while 
    for GANs it corresponds to a saddle point that is a local maximum for the classifier and a local
    minimum for the generator. Asymptotic consistency is proven in the space of unconstrained
    probability distribution functions; when we move to specific parametric families of distributions
    a variety of caveats apply: the given function family may not include the true training data distribution,
    different parameter values may encode the same function and thus introduce identifiability issues, and
    the optimizer may fail to find the global optimum.

    Because GANs and NCE both use the same value function, it is natural to wonder how they are related
    to each other, and to maximum likelihood. In this paper, we provide some initial answers to each
    question. We show

\begin{itemize}
    \item A modified version of NCE with a dynamic generator is equivalent to MLE.
    \item The existing theoretical work on GANs does not guarantee convergence on
        practical applications.
    \item Because GANs do the model estimation in the generator network, they
        cannot recover maximum likelihood using $V$.
\end{itemize}

Throughout this paper, we make some weak regularity assumptions.
Specifically, we will assume that all of our models, be they $p_m$ or $p_g$, parameterize
the probability distribution such that
$p(\vx) > 0$ for all $\vx$.
In the case of continuous random variables,
we additionally assume that
that $p$ and $\nabla_{\vtheta} p$ are continuous at all $\vx$ and $\vtheta$ points.

\section{Self-contrastive estimation}

The performance of NCE is highly dependent on the choice of noise distribution.
It is not difficult to discriminate data samples from totally unstructured noise,
so models trained with too simplistic of a noise distribution often underfit badly.
This has
motivated a variety of heuristic attempts to design better noise distributions.
~\citet{nce} suggest
that ``one could choose a noise distribution by first estimating a preliminary model
of the data, and then use this preliminary model as the noise distribution.''

Let use consider the extreme version of this approach, where the model is copied and used
as the new noise distribution after every step of learning. We call this approach
{\em self-contrastive estimation}. Here we show that self-contrastive estimation has the same expected gradient as
maximum likelihood estimation.

Let $\theta$ be a parameter of $p_m$. Then
\begin{equation}
    \frac{\partial}{\partial \theta} V(p_c, p_g)  = \frac{\partial}{\partial \theta}
    \left[
\mathbb{E}_{\vx \sim p_d} \log p_c(y=1 \mid \vx) +
\mathbb{E}_{\vx \sim p_g} \log p_c(y=0 \mid \vx)
\right]
\label{eq:foo}
\end{equation}

Recall from Eq.~\ref{eq:nce_c} that in NCE, the classifier is defined by
\[ p_c(y= 1\ \mid \vx) = \frac{p_m(\vx)} {p_m(\vx) + p_g(\vx) }. \]
In the context of SCE, we have copied $p_m$ into $p_g$ before each step of learning.
It may therefore be tempting to make the assumption that $p_m = p_g$ and simplify
the classifier to $p_c(y = 1 \mid \vx) = \frac{1}{2}$. However, this is incorrect,
because we have made a deep copy, rather than aliasing $p_g$ to $p_m$. It is crucial
that $p_g$ is not dependent on $\theta$; therefore when calculating derivatives
we must consider the effect of $\theta$ on $p_m$ but not $p_g$.
Substituting Eq.~\ref{eq:nce_c} into Eq.~\ref{eq:foo} we obtain
\[
    \frac{\partial}{\partial \theta} V(p_c, p_g)  = \frac{\partial}{\partial \theta}
    \left[
    \mathbb{E}_{\vx \sim p_d} \log \frac{p_m(\vx)}{p_m(\vx)+p_g(\vx)} +
    \mathbb{E}_{\vx \sim p_g} \log \frac{p_g(\vx)}{p_m(\vx)+p_g(\vx)}
\right]
\]
\[
\ = \frac{\partial}{\partial \theta}
\left[
\mathbb{E}_{\vx \sim p_d} \left[
    \log p_m(\vx) - \log\left( p_m(\vx)+p_g(\vx) \right)
\right]+
\mathbb{E}_{\vx \sim p_g} \left[
    \log p_g(\vx) - \log\left(p_m(\vx)+p_g(\vx)\right)
    \right]
\right]
\]
\[
 = \frac{\partial}{\partial \theta}
\left[
\mathbb{E}_{\vx \sim p_d} \left[
    \log p_m(\vx) - \log\left( p_m(\vx)+p_g(\vx) \right)
\right]
- \mathbb{E}_{\vx \sim p_g}  \log\left(p_m(\vx)+p_g(\vx)\right)
\right]
\]
\begin{equation}
 =
\mathbb{E}_{\vx \sim p_d}
\frac{\partial}{\partial \theta}
\left[
    \log p_m(\vx) - \log\left( p_m(\vx)+p_g(\vx) \right)
\right]
-
\mathbb{E}_{\vx \sim p_g}
\frac{\partial}{\partial \theta}
\log\left(p_m(\vx)+p_g(\vx)\right)
\label{eq:bar}
\end{equation}

We can show that the term on the right vanishes, as follows.
\[
\mathbb{E}_{\vx \sim p_g}
\frac{\partial}{\partial \theta}
\log\left(p_m(\vx)+p_g(\vx)\right)
\]
\[
=
\mathbb{E}_{\vx \sim p_g}
\frac{
    \frac{\partial}{\partial \theta}
    p_m(\vx)
}
{
p_m(\vx)+p_g(\vx)
}.
\]
Because we have assumed our distributions are strictly positive, we
can use the trick $\frac{\partial}{\partial \theta} p(\vx)
= \frac{\partial}{\partial \theta} \exp \left( \log p(\vx) \right) 
= p(\vx) \frac{\partial}{\partial \theta} \log p(\vx)$:
\[
\mathbb{E}_{\vx \sim p_g}
\frac{
    \frac{\partial}{\partial \theta}
    p_m(\vx)
}
{
p_m(\vx)+p_g(\vx)
}
=
\mathbb{E}_{\vx \sim p_g}
\frac{
    p_m(\vx) 
    \frac{\partial}{\partial \theta}
    \log p_m(\vx)
}
{
p_m(\vx)+p_g(\vx)
}.
\]
Crucially, outside of the differentiation sign, we {\em are} allowed to exploit the
fact that $p_m(\vx)$ and $p_g(\vx)$ are equal in order to simplify $p_m / (p_m + p_g)$ into $\frac{1}{2}$.
We are left with
\[
    \frac{1}{2} \mathbb{E}_{\vx \sim p_g} \log p_g(\vx).
\]
We can use the inverse of the $\exp(\log p(\vx))$ trick to observe that this is equal to
$\frac{\partial} {\partial \theta} \sum_\vx p_g(\vx)
= \frac{\partial} {\partial \theta} 1 = 0$. (In the continuous case one must use an integral
rather than a summation)

Plugging this result in Eq.~\ref{eq:bar}, we obtain:
\[
    \frac{\partial}{\partial \theta} V(p_c, p_g)  = 
\mathbb{E}_{\vx \sim p_d}
\frac{\partial}{\partial \theta}
\left[
    \log p_m(\vx) - \log\left( p_m(\vx)+p_g(\vx) \right)
\right]
\]
Using the same tricks as previously, we can simplify this to
\[
    \frac{1}{2} \mathbb{E}_{\vx \sim p_d}
\frac{\partial}{\partial \theta}
\log p_m(\vx).
\]
This is $\frac{1}{2}$ the log likelihood gradient, and the $\frac{1}{2}$ can of course be folded
into the learning rate.

Note that while the gradients are equivalent, the objective functions are not.
MLE maximizes a single objective function, while SCE changes the objective function at every step.
The MLE gradient for a specific value of $\bm{\theta}$ always matches the gradient at $\bm{\theta}$
of a different SCE objective constructed specifically for that $\bm{\theta}$. The value of the
SCE objective does not change over time; each new objective function always has value $-2\log 2$ as
a consequence of the model distribution never being distinguishable from the noise distribution.

\section{Interpreting the theory of GANs}

One can prove the asymptotic consistency of MLE, NCE, and GANs by observing that
each minimizes a convex divergence in function space. In practice, one cannot
optimize directly over probability distribution functions. Instead, one must optimize
over parameters indexing a parametric family of functions. Fortunately, estimators
that are consistent in function space often prove to perform well in parameter space.

In the case of GANs, a subtlety of the theory may have important implications for
the behavior of the algorithm in practice. The existing theory justifying GANs uses
convexity in function space to prove not only asymptotic consistency but also convergence.

Estimating a model by maximimizing the MLE or NCE objective function is guaranteed to
converge for smooth functions that are bounded from above regardless of whether these
objective functions are convex. It is possible for optimization to get stuck in a local
maximum in parameter space, but the optimization process will at least arrive at some
critical point.

In the case of GANs, the generator is updated while holding the discriminator fixed,
and vice versa. In function space this corresponds to performing subgradient descent
on a convex problem, and is guaranteed to converge. 

In the non-convex case, the existing theory does not specify what will happen.

To reach the equilibrium point, $p_g$ should be trained to minimize
    $\max_{p_c} V(p_c, p_d)$. Instead, it takes successive steps partially
    minimizing $V(p_c, p_d)$ using the current value of $p_c$ at each step.
    Because by definition $V(p_c, p_d) \leq \max_{p_c} V(p_c, p_d)$ this
    corresponds to taking steps that partially {\em minimize a lower bound}.
    In the non-convex case, it is conceivable that this could merely make
    the bound looser rather than decrease the underlying objective function
    as desired. This may result in the learning procedure oscillating rather
    than converging.
    
    Such a process could explain the underfitting that has
    been observed with GANs thus far. No deep generative model has been
    demonstrated to be able to memorize a rich, complicated training set.
    For many models, this could be explained by inaccuracies in the approximation
    of the gradient or too strong of simplifying assumptions for variational
    learning. For GANs, the failure to memorize the training set is surprising
    because the gradient can be computed with backpropagation and there are no
    variational approximations. Non-convergence of gradient-based learning
    for continuous games stands out as a candidate explanation for why this happens.

    To be clear, we do not have any positive results identifying non-convergence as the problem.
    We are merely identifying one way in which the existing theoretical results for
    GANs {\em fail to guarantee} good performance in practice. We suggest that future
    work could attempt to positively identify non-convergence in GAN learning or
    apply better algorithms for computing the equilibrium of the game.

\section{GANs cannot implement maximum likelihood}

GANs work by learning in the {\em generator}, while NCE works by learning in the discriminator
(via a generative model that is used to implicitly define the generator).
This turns out to have important results for learning. Namely, each step of learning in a GAN pair consists
of decreasing an expectation of a function of samples from the generator:
\[
\mathbb{E}_{\vx \sim p_g} f(\vx)
\]
where $f(\vx) = \log p_c\left(y=0 \mid \vx \right)$.

For a parameter $\theta$ of $p_g$, we find
\[
\frac{\partial}{\partial \theta} \mathbb{E}_{\vx \sim p_g} f(\vx) 
\]
\[
    = \int f(\vx) \frac{\partial}{\partial \theta} p_g(\vx)
\]
\[
    = \int f(\vx) p_g(\vx) \frac{\partial} {\partial \theta} \log p_g(\vx).
\]

\begin{figure}
  \centering
  \includegraphics[width=\textwidth]{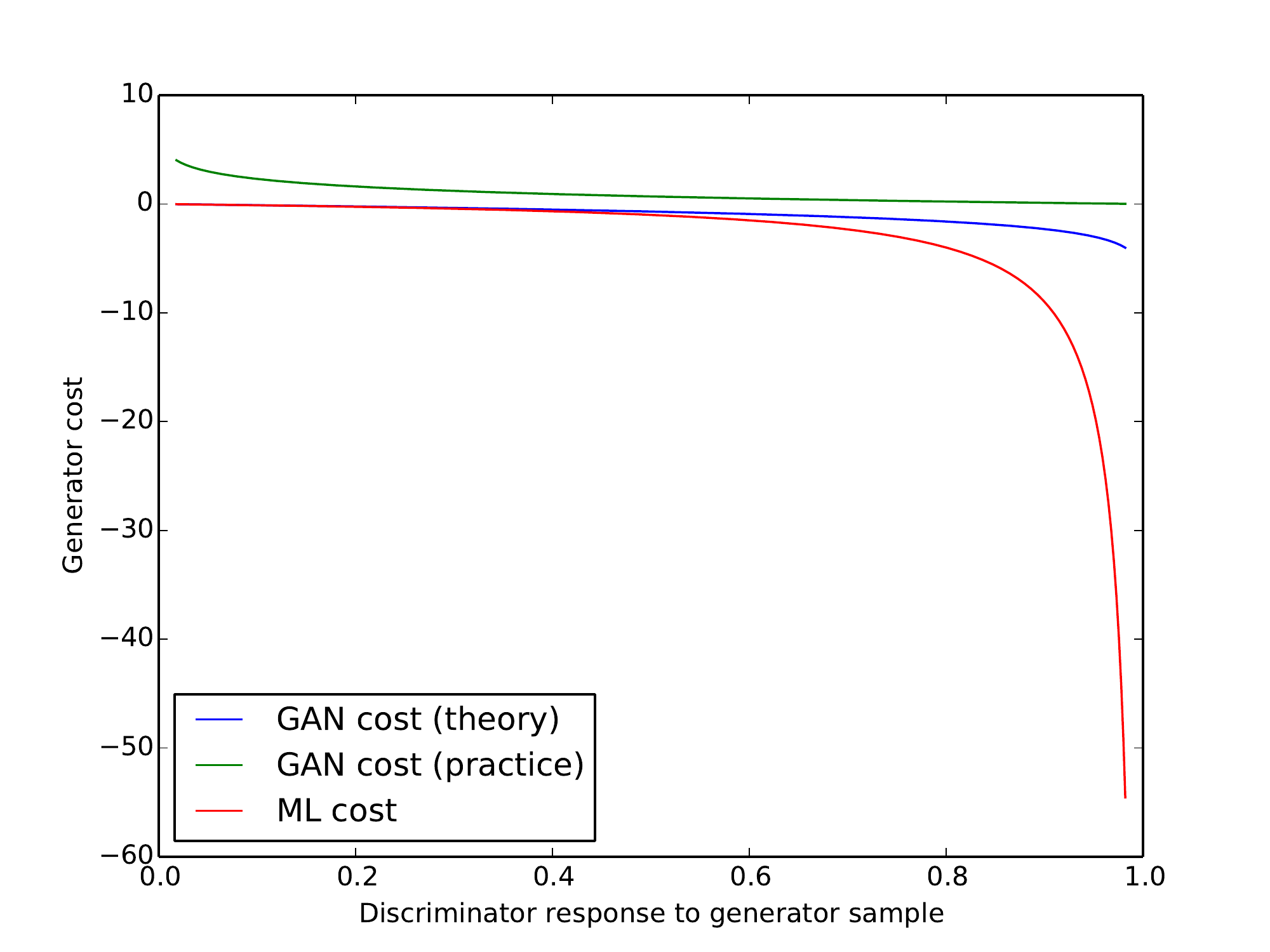}
  \caption{The cost the generator pays for sampling a point is a function of the optimal discriminator's
    output for that point. This much is true for maximum likelihood, the minimax formulation
    of the distinguishability game, and a heuristic reformulation used in most experiments by
    ~\citet{gan}. Where the methods differ is the exact value of that cost. As we can see, the
    cost for maximum likelihood only has significant gradient through the discriminator if the
    discriminator is fooled with high confidence. Since this is an extremely rare event when sampling
    from an untrained generator, estimates
    of the maximum likelihood gradient based on this approach have high variance.
  }
  \label{fig:cost_diff}
\end{figure}

From this vantage point it is clear that to obtain the maximum likelihood derivatives, we need
\[ 
    f(\vx) = - \frac{p_d(\vx)} {p_g(\vx)}.
\]
(We could also add an arbitrary constant and still obtain the correct result)
Suppose our discriminator is given by $p_c(y=1 \mid \vx) = \sigma\left( a(\vx) \right)$ where $\sigma$ is the
logistic sigmoid function. Suppose further that our discriminator has converged to its optimal value for
the current generator, 
\[ p_c(y = 1 \mid \vx) = \frac{ p_d(\vx) } {p_g(\vx) + p_d(\vx)}.\]
Then $f(x) = - \exp\left( a( \vx ) \right)$. 
This is clearly different from the value given by the distinguishability game, which simplifies to
$f(\vx) = -\zeta\left( a( \vx \right))$, where $\zeta$ is the softplus function. See this function
plotted alongside the GAN cost in Fig~\ref{fig:cost_diff}.

In other words, the discriminator gives us the necessary information to compute the maximum likelihood
gradient of the generator, but it requires that we abandon the distinguishability game.
In practice, the estimator based on $\exp \left( a(\vx) \right)$ has too high of variance.
For an untrained model, sampling from the generator almost always yields very low values of $\frac{p_d(\vx)}{p_g(\vx)}.$
The value of the expectation is dominated by the rare cases where the generator manages to sample
something that resembles the data by chance. Empirically, GANs have been able to overcome this problem but
it is not entirely clear why. Further study is needed to understand exactly what tradeoff GANs are
making.

\section{Discussion}

Our analysis has shown a close relationship between noise contrastive estimation and maximum likelihood.
We can now interpret noise-contrastive estimation as being a one-sided version of a distinguishability
game.
Our analysis has also shown
that generative adversarial networks are not as closely related to noise contrastive estimation as
previously believed. The fact that the primary model is the generator turns out to result in a departure
from maximum likelihood even in situations where NCE and MLE are equivalent. Further study is needed to
understand exactly what tradeoff is incurred when using adversarial learning.
Finally, the problem of non-convergence of independent SGD in
the non-convex case may explain the underfitting observed in GANs and suggests the application of better
algorithms for solving for the equilibrium strategies of the distinguishability game.

\subsubsection*{Acknowledgments}
We would like to thank Andrew Dai, Greg Corrado, David Sussillo, Jeff Dean
and David Warde-Farley
for their feedback on drafts of this article.

\small
\setlength{\bibsep}{5pt plus 0.3ex}
\bibliography{iclr2015}
\bibliographystyle{iclr2015}
\end{document}